\documentclass{article}

     \PassOptionsToPackage{numbers, compress}{natbib}

\usepackage[preprint]{neurips_2019}
\usepackage{algorithmic,algorithm}
\usepackage{wrapfig,sidecap}
\usepackage{subcaption}

\usepackage{enumitem}

\usepackage{amsmath}
\usepackage{amsthm}
\theoremstyle{definition}
\newtheorem{definition}{Definition}[section]
\usepackage{comment}
\usepackage{bm}
\DeclareMathAlphabet\mathbfcal{OMS}{cmsy}{b}{n}
\newcommand{\ten}[1]{\mathbfcal{#1}} 
\newcommand{\mat}[1]{\mathbf{#1}}

\usepackage{tikz}
\usetikzlibrary{bayesnet}
\usetikzlibrary{shapes, arrows, calc, positioning,matrix}
\tikzset{
data/.style={circle, draw, text centered, minimum height=3em ,minimum width = .5em, inner sep = 2pt},
empty/.style={circle, text centered, minimum height=3em ,minimum width = .5em, inner sep = 2pt},
}

\usepackage[utf8]{inputenc} 
\usepackage[T1]{fontenc}    
\usepackage{hyperref}       
\usepackage{url}            
\usepackage{booktabs}       
\usepackage{amsfonts}       
\usepackage{nicefrac}       
\usepackage{microtype}      

\title{Bayesian Tensorized Neural Networks with Automatic Rank Selection}

\author{%
  Cole ~Hawkins \\
  Department of Mathematics\\ University of California, Santa Barbara\\
  \texttt{colehawkins@math.ucsb.edu} \\
   \And
   Zheng Zhang \\
   Department of ECE\\ University of California, Santa Barbara\\
   \texttt{zhengzhang@ece.ucsb.edu} \\
}

\begin{document}

\maketitle

\begin{abstract}
Tensor decomposition is an effective approach to compress over-parameterized neural networks and to enable their deployment on resource-constrained hardware platforms. However, directly applying tensor compression in the training process is a challenging task due to the difficulty of choosing a proper tensor rank. In order to address this challenge, this paper proposes a low-rank Bayesian tensorized neural network. Our Bayesian method performs automatic model compression via an adaptive tensor rank determination. We also present approaches for posterior density calculation and maximum a posteriori (MAP) estimation for the end-to-end training of our tensorized neural network. We provide experimental validation on a two-layer fully connected neural network, a 6-layer CNN and a 110-layer residual neural network where our work produces $7.4\times$ to $137\times$ more compact neural networks directly from the training while achieving high prediction accuracy.
\end{abstract}

\section{Introduction}
Despite their success in many engineering applications, deep neural networks are often over-parameterized, requiring extensive computing resources in their training and inference. 
In order to deploy neural networks on resource-constrained platforms such as IoT devices and smart phones, numerous techniques have been developed to build {\it compact} neural network models~\citep{alvarez2017compression,hanson1989comparing,lecun1990optimal,lebedev2014speeding}. 

As a high-order generalization of matrix factorization, tensor decomposition has outperformed many existing compression algorithms by exploiting the hidden low-rank structures in high dimensions. Given an over-parameterized neural network, such techniques have achieved state-of-the-art performance by tensorizing and compressing the fully connected layers or convolution kernels~\citep{lebedev2014speeding,garipov2016ultimate,kossaifi2017tensor,novikov2015tensorizing,yu2017long,kim2015compression}. However, the train-then-compress framework cannot avoid the prohibitive computational costs in the training process.

This paper tries to directly train a low-rank tensorized neural network. A main challenge is to automatically determine the tensor rank (and thus the model complexity). Exactly determining a tensor rank is NP-hard~\cite{hillar2013most}. Existing tensor rank surrogates~\cite{gandy2011tensor,lu2019tensor} are not suitable for regularizing the training process due to their high computational costs. Meanwhile, the unknown tensors describing weight matrices and convolution kernels are embedded in a nonlinear learning model. Although~\cite{novikov2015tensorizing,calvi2019tucker} presented some approaches to train tensorized neural networks, they assume that the tensor ranks are given, which is often infeasible in practice. Recently, some methods have been reported to control the rank of tensor trains in some optimization problems~\cite{dolgov2014alternating,holtz2012alternating}, but they often cause over-fitting because the tensor ranks can be arbitrarily increased in order to minimize the objective functions.


{\bf Contributions.} Inspired by the recent Bayesian CP and Tucker tensor completion~\cite{zhao2015bayesian,zhao2016bayesian}, we develop a novel low-rank Bayesian tensorized neural network. Our contribution is two-fold. Firstly, we present a Bayesian model to compress the model parameters (e.g., weight matrices and convolution kernels) via tensor train decomposition~\cite{oseledets2011tensor}. Our method employs a proper prior density to automatically determine the tensor ranks based on the given training data, which is beyond the capability of existing tensorized neural networks. Secondly, we develop training algorithms to estimate the full posterior density and the MAP point. To solve the large-scale Bayesian inference, we approximate the posterior density by Stein variational gradient descent~\cite{liu2016stein}. The proposed framework can generate much more compact neural networks. Our method can also provide uncertainty estimation, which is important for certain applications like decision making in autonomous driving~\cite{kendall2017uncertainties} and robotics. 

\subsection{Related Work}

Our work is related to and based on the following research results. 

{\bf Tensor Compression of Neural Networks.}
CP-format tensor decomposition~\cite{carroll1970analysis,harshman1994parafac} was first employed to compress the fully connected (FC) layers of pre-trained models by~\citet{lebedev2014speeding}. Following work then compressed both FC and convolutional layers by tensor train decomposition~\cite{garipov2016ultimate} and by Tucker decomposition~\cite{kim2015compression}, respectively. The results in~\cite{garipov2016ultimate,kim2015compression} show that the FC layers can be compressed more significantly than the convolution layers. In order to avoid the expensive pre-training, \citet{novikov2015tensorizing} and \citet{calvi2019tucker} trained FC layers in low-rank tensor-train and Tucker formats, respectively, with the tensor ranks fixed in advance.  In practice determining a proper tensor rank {\it a priori} is hard, and a bad rank estimation can result in low accuracy or high training cost. This challenge is the main motivation of our work.

{\bf Tensor Rank Determination.} Two main approaches are used for rank determination in tensor completion: low-rank optimization and Bayesian inference. Optimization methods mainly rely on some generalization of the matrix nuclear norm~\cite{recht2010guaranteed} to tensors. The most popular approaches place a low-rank objective on tensor unfoldings~\cite{gandy2011tensor,liu2013tensor,imaizumi2017tensor}, but the computation is expensive for high-order tensors. The Frobenius norms of CP factors are used as a regularization for 3-way tensor completion~\cite{bazerque2013rank}, but this does not generalize to high-order tensors either. Bayesian methods can directly infer the tensor rank in CP or Tucker tensor completion through low-rank priors~\cite{zhao2015bayesian,rai2014scalable,guhaniyogi2017bayesian,bazerque2013rank,hawkins2018robust}. In the CP and Tucker formats, the ranks of different tensor factors do not couple with each other. This is not true in the tensor-train format (see Section~\ref{sec: lr_prior}). Furthermore, in tensor completion the observed data is a linear mapping of a tensor, whereas the mapping is nonlinear in neural networks. Therefore, previous work on tensor rank determination cannot be directly applied to tensorized neural networks.

{\bf Bayesian Neural Networks.} Bayesian neural networks were developed to quantify the model uncertainty~\cite{neal1995bayesian}. Both~\citet{mackay1992bayesian} and~\citet{neal1995bayesian} explored sparsity-inducing shrinkage priors for relevance determination and weight removal in neural networks. Another popular prior for Bayesian neural network compression is the Minimum Description Length (MDL) framework for quantization~\cite{hinton1993keeping}. MDL has seen modern implementations by~\citet{ullrich2017soft} and~\citet{louizos2017bayesian}, who also investigated Bayesian pruning. To our best knowledge, no Bayesian approaches have been reported to train a low-rank tensorized neural network. A main challenge of training a Bayesian neural network is the high computational cost caused by sampling-based posterior density estimations. We will address this issue by the recently developed Stein variational gradient descent~\cite{liu2016stein}. 

\section{Background: Tensor Train (TT) Decomposition}

\paragraph{Notations.} We use bold lowercase letters (e.g., $\mat{a}$), uppercase letters (e.g., $\mat{A}$) and bold calligraphic letters (e.g., $\ten{A}$) to represent vectors, matrices and tensors, respectively. A tensor is a generalization of a matrix, or a multi-way data array~\cite{kolda2009tensor}. An order-$d$ tensor is a $d$-way data array $\ten{A}\in \mathbb{R}^{I_1 \times I_2 \times \dots \times I_d}$, where $I_k$ is the size of mode $k$. Given the integer $i_k \in [1, I_k] $ for each mode $k=1 \cdots d$, an entry of the tensor $\ten{A}$ is denoted by $\ten{A}(i_1,\cdots, i_d)$. A {\bf subtensor} is obtained by fixing a subset of the tensor indices. A {\bf slice} is a two-dimensional section of a tensor, obtained by fixing all but two
indices. 



\begin{definition}\label{def: tt-format}
The {\bf tensor-train} (TT) factorization~\cite{oseledets2011tensor} uses a compact multilinear format to express a tensor $\ten{A}$. Specifically, it expresses a $d$-way tensor $\ten{A}$ as a collection of matrix products:
\[
\ten{A}(i_1,i_2,\dots,i_d) = \ten{G}_1(:,i_1,:)\ten{G}_2(:,i_2,:)\dots\ten{G}_d(:,i_d,:)
\]
where $\ten{G}_k\in \mathbb{R}^{R_{k-1}\times I_k \times R_{k}}$ and $R_0=R_d=1$. The vector $\mat{R}=(R_0,R_1,R_2,\dots,R_d)$ is the {\bf TT-rank}. Each $3$-way tensor $\ten{G}_i$ is a {\bf TT core}.
\end{definition}



The tensor-train decomposition can also be applied to compress a matrix $\mat{W}\in \mathbb{R}^{M\times J}$. For the following definition we assume that the dimensions $M$ and $J$ can be factored as follows:
\begin{equation}
\label{eq:mat2ten dimensions}
M=\prod_{k=1}^d M_k \hspace{.2in} J=\prod_{k=1}^d J_k. \nonumber
\end{equation}
Let $\mu,\nu$ be the natural bijections from indices $(m,j)$ of $\mat{W}$ to indices $(\mu_1(m),\nu_1(j),\dots,\mu_d(m),\nu_d(j))$ of an order-$2d$ tensor $\ten{W}$. We identify the entries of $\mat{W}$ and $\ten{W}$ in the natural way:
\begin{equation}
\label{eq:mat2ten}
\mat{W}(m,j) = \ten{W}(\mu_1(m),\nu_1(j),\dots,\mu_d(m),\nu_d(j)).
\end{equation}

\begin{definition}
The {\bf TT-matrix} factorization expresses the matrix $\mat{W}$ as a series of matrix products:
\begin{equation}
\label{eq:tt_mat}
\ten{W}(\mu_1(m),\nu_1(j),\dots,\mu_d(m),\nu_d(j)) = \prod \limits_{k=1}^d \ten{G}_k\left(:,\mu_k\left( m \right),\nu_k\left( j\right),:\right)
\end{equation}
where each $4$-way tensor $\ten{G}_k\in \mathbb{R}^{R_{k-1}\times M_k\times J_k \times R_{k}}$ is a TT core, and $R_0=R_{d}=1$. The vector $\mat{R}=(R_0,R_1,\dots,R_{d})$ is again the TT-rank. 
\end{definition}
For notational convenience we express a TT or TT-matrix parameterization of $\ten{W}$ as 
\begin{equation}
    \ten{W}= [\![ \ten{G}_1,\dots,\ten{G}_d]\!].
\end{equation}
The TT-matrix format reduces the total number of parameters from $\prod_{k=1}^d M_k\prod_{k=1}^d J_k$ to $\sum_{k=1}^d R_{k-1}M_k J_k R_{k}$. This leads to massive parameter reductions when the TT-rank is low.

\section{Our Proposed Method}

\subsection{Low-Rank Bayesian Tensorized Neural Networks (LR-BTNN)}
Let $\mathcal{D} = \{(\mat{x}_i,\mat{y}_i)\}_{i=1}^N$ be the training data, where $\mat{y}_i \in \mathbb{R}^S$ is an output label. We intend to train an $L$-layer tensorized neural network
\begin{equation}
   \mat{y} \approx \mat{g}\left( \mat{x} | \; \{ \ten{W}^{(l)}\}_{l=1}^L \right).
\end{equation} 
Here $\ten{W}^{(l)}$ is an unknown tensor describing the massive model parameters in the $l$-th layer. We consider two kinds of tensorized layers in this paper:
\begin{itemize}[leftmargin=*]
\item {\bf TT-FC Layer.} In a fully connected (FC)  layer, a vector is mapped to a component-wise nonlinear activation function by a weight matrix $\mat{W}^{(l)}$. We tensorize $\mat{W}^{(l)}$ as $\ten{W}^{(l)}$ according to \eqref{eq:mat2ten}.

\item {\bf TT-Conv Layer.} A convolutional filter takes the form $\ten{K}^{(l)}\in\mathbb{R}^{t\times t \times C \times S}$ where $t\times t$ is the filter size, $C$ and $S$ are the numbers of input and output channels, respectively. The number of channels $C$ and $S$ are often larger than the filter size $t$, so we factorize $C=\prod_{i=1}^d c_i, S=\prod_{i=1}^d s_i$, reshape $\ten{K}^{(l)}$ into a $t^2\times c_1\times \dots \times c_d\times s_1\times \dots \times s_d$ tensor, and denote the reshaped tensor as $\ten{W}^{(l)}$. 

\end{itemize}

Our goal is to parameterize and compress each unknown $\ten{W}^{(l)}$ with a TT-matrix format in the training process. To simplify notations, we consider a single-layer neural network parametrized by a single tensor $\ten{W}$, but our results can be easily generalized to a general $L$-layer network (see our result section). In order to build a Bayesian model, we assume the following likelihood function
\begin{equation}
    \label{likelihood}
    p\left(\ten{D}| \left\{ \ten{G}_k \right \}_{k=1}^d \right) = \prod_{i=1}^N p\left(\mat{y}_i,\mat{g}\left(\mat{x}_i \mid [\![ \ten{G}_1,\dots,\ten{G}_d]\!]  \right) \right).
\end{equation}
Here $ \ten{G}_{k} \in \mathbb{R}^{R_{k-1}\times M_k \times J_k \times R_k}$ is an unknown TT core of $\ten{W}$, and $\mat{R}=(1, R_1, \dots, R_{d-1},1)$ sets an {\bf upper bound} for the TT rank. The actual TT rank will be determined later. In this paper we focus on classification problems so we assume a multinomial likelihood. Let $y_{i,s}$ and $g_s$ be the $s$-th component of $\mat{y}_i$ and $\mat{g}$, respectively, then we have 
\begin{equation}
    \label{eq: multinomial likelihood}
p\left(\mat{y}_i,\mat{g}\left(\mat{x}_i \mid [\![ \ten{G}_1,\dots,\ten{G}_d]\!]  \right) \right) = \prod_{s=1}^S g_s\left(\mat{x}_i \mid [\![ \ten{G}_1,\dots,\ten{G}_d]\!]  \right)^{y_{i,s}}.
\end{equation}
As a result, the negative log-likelihood is the standard cross-entropy loss
\begin{equation}
    \mathcal{L}(\{\ten{G}_k\})=-\sum_{i=1}^N \sum_{s=1}^S y_{i,s}\log g_s\left(\mat{x}_i \mid [\![ \ten{G}_1,\dots,\ten{G}_d]\!]  \right).
\end{equation}

In order to infer the unknown TT cores $\{ \ten{G}_k\}_{k=1}^d$ and to decide the actual ranks, we will further introduce some hidden variables $\{ \boldsymbol{\lambda}^{(k)} \}_{k=1}^{d-1}$ to parameterize the prior density of $\{ \ten{G}_k\}_{k=1}^d$ (which will be explained in Section~\ref{sec: lr_prior}).
Let $\boldsymbol{\theta}$ denote all unknown variables 
\begin{equation}
    \boldsymbol{\theta}:=\left \{ \left\{ \ten{G}_k \right \}_{k=1}^d, \{ \boldsymbol{\lambda}^{(k)} \}_{k=1}^{d-1} \right \}
\end{equation}
which is described with a prior density $p(\boldsymbol{\theta})$. Then we can build a tensorized neural network by estimating the MAP point or the full distribution of the following posterior density function:
\begin{equation}
    \label{eq: posterior}
    p(\boldsymbol{\theta} |\ten{D}) =  \frac{p(\ten{D}|\boldsymbol{\theta})p(\boldsymbol{\theta})}{p(\ten{D})} 	\propto  
   p(\ten{D}|\boldsymbol{\theta})p(\boldsymbol{\theta})=p(\ten{D},\boldsymbol{\theta}).
\end{equation} 
There exist two key challenges. Firstly, how shall we choose the prior density $p(\boldsymbol{\theta})$ to ensure desired model structures? Secondly, how can we efficiently solve the resulting large-scale Bayesian inference?


\subsection{Selection of Prior Density Functions}
\label{sec: lr_prior}
\begin{figure}[t]
\centering
\begin{minipage}{.35\linewidth}
  \resizebox{\textwidth}{!}{
  \begin{tikzpicture}
    \matrix[matrix of nodes,nodes={draw},column 1/.style={nodes={fill=blue!10}},column 2/.style={nodes={fill=green!30}},column 3/.style={nodes={fill=red!45}},every node/.style={yscale=0.95}] {
      $\ten{G}_k(1,i_k,1)$ & $\ten{G}_k(1,i_k,2)$ & $\ten{G}_k(1,i_k,3)$ \\
      $\ten{G}_k(2,i_k,1)$ & $\ten{G}_k(2,i_k,2)$ & $\ten{G}_k(2,i_k,3)$ \\
    };
\end{tikzpicture}
    }
\centering  \small { Slice of TT-core $\ten{G}_{k}$}
  \end{minipage}
 \begin{minipage}{.55\linewidth}
  \resizebox{\textwidth}{!}{
    \begin{tikzpicture}

    \matrix[thick,matrix of nodes,nodes={draw},row 1/.style={nodes={fill=blue!10}},row 2/.style={nodes={fill=green!30}},row 3/.style={nodes={fill=red!45}},every node/.style={xscale=0.85}] {
 $\ten{G}_{k+1}(1,i_{k+1},1)$ & $\ten{G}_{k+1}(1,i_{k+1},2)$ & $\ten{G}_{k+1}(1,i_{k+1},3)$  & $\ten{G}_{k+1}(1,i_{k+1},4)$ \\
      $\ten{G}_{k+1}(2,i_{k+1},1)$ & $\ten{G}_{k+1}(2,i_{k+1},2)$ & $\ten{G}_{k+1}(2,i_{k+1},3)$&$\ten{G}_{k+1}(2,i_{k+1},4)$ \\
      $\ten{G}_{k+1}(3,i_{k+1},1)$ & $\ten{G}_{k+1}(3,i_{k+1},2)$ & $\ten{G}_{k+1}(3,i_{k+1},3)$ &$\ten{G}_{k+1}(3,i_{k+1},4)$\\
    };
  \end{tikzpicture}
  }
\centering \small \\ { Slice of TT-core $\ten{G}_{k+1}$}
 \end{minipage}
 \caption{Elements of $3$-way $\ten{G}_k$ and $\ten{G}_{k+1}$. The elements controlled by the same entry of $\boldsymbol\lambda^{(k)}$ are marked with the same color. Here the upper bound of TT rank is set as  $R_{k-1}=2,R_k=3,R_{k+1}=4$.}
 \label{fig:tt rank det}
\end{figure}
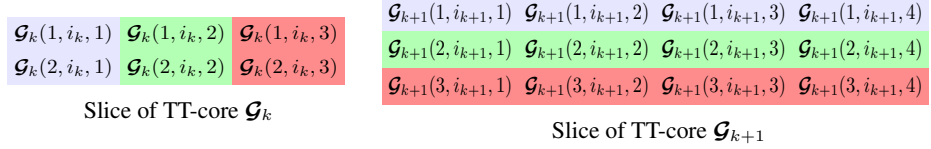

In order to achieve automatic model compression, the prior density function in \eqref{eq: posterior} should be chosen such that: (1) the actual rank of $ \ten{G}_{k}$ is very small; (2) manual rank tuning can be avoided. 

{\bf Prior Density for $\ten{G}_k$.} We specify the prior density on a TT-matrix core. The size of $\ten{G}_k$ is fixed as $R_{k-1} \times M_k \times J_k \times R_k$. In order to reduce the TT rank, we will enforce some rows and columns in the slice $\ten{G}_k (:, m_k, j_k, :)$ to zero.
The main challenge is that the matrix products are coupled: the $r_k$-th column of $\ten{G}_{k-1} (:, m_{k-1}, j_{k-1},:)$ and the $r_k$-th row of $\ten{G}_{k} (:, m_{k}, j_{k},:)$ should simultaneously shrink to zero if a rank deficiency happens. Fig.~\ref{fig:tt rank det} uses the slices of two adjacent 3-way TT cores to show this coupling in the TT decomposition. 

In order to address the above challenge, we extend the method in~\cite{zhao2015bayesian, zhao2016bayesian} which was developed for CP and Tucker tensor completion but was not applicable for neural networks or TT format. Specifically, we introduce the non-negative vector parameter $\boldsymbol{\lambda}^{(k)} =[\lambda_1^{(k)}, \dots, \lambda_{R_k}^{(k)}]\in \mathbb{R}^{R_k}$ to control the actual rank $\hat{R}_k$ for each $k$ with $1\leq k\leq d-1$.  For each intermediate TT core $\ten{G}_k$, we place a normal prior on its entries:
\begin{equation}
    \label{eq:core tensor column prior}
    p\left(\ten{G}_k\:|\:\boldsymbol{\lambda}^{(k-1)},\boldsymbol{\lambda}^{(k)}\right) = \prod_{r_{k-1},m_k,j_k,r_k} \mathcal{N}\left(\ten{G}_k(r_{k-1},m_k,j_k,r_k)\:|\: 0,\lambda_{r_{k-1}}^{(k-1)}\cdot \lambda_{r_k}^{(k)} \right), \; 2\leq k \leq d-1.
\end{equation}
where $1\leq r_{k-1}\leq R_{k-1},1\leq r_k\leq R_{k}, 1\leq m_k\leq M_k$ and $1\leq j_k \leq J_k$. 

Because $R_0=R_d=1$, the ranks of the first and last TT cores are controlled by only one vector. We place a similar normal prior on each: 
\begin{align}
\small
    \label{eq: first core prior}
      \begin{split}
        p\left(\ten{G}_1\:|\:\boldsymbol{\lambda}^{(1)}\right) &= \prod_{m_1, j_1, r_1} \mathcal{N}\left(\ten{G}_1(1,m_1, j_1,r_1)\:|\: 0,\left(\lambda_{r_1}^{(1)}\right)^2 \right)\\
        p\left(\ten{G}_d\:|\:\boldsymbol{\lambda}^{(d-1)}\right) &= \prod_{r_{d-1},m_d,j_d} \mathcal{N}\left(\ten{G}_{d}(r_{d-1},m_d,j_d,1)\:|\: 0,\left(\lambda_{r_{d-1}}^{(d-1)}\right)^2 \right).\\
      \end{split}
      \normalsize
\end{align}
Here we use the squares $\left(\lambda_{r_1}^{(1)}\right)^2$ and $\left(\lambda_{r_{d=1}}^{(d-1)}\right)^2$ to ensure that the order of magnitude of the priors is consistent across all TT cores. To apply the same prior to the TT cores of the standard tensor train we remove the third index $j_k$ of each TT core.

{\bf Prior Density for $\boldsymbol\lambda^{(k)}$.} In order to avoid tuning $\{ \boldsymbol\lambda^{(k)} \}$ and TT-ranks manually, we set each $\boldsymbol\lambda^{(k)}$ as a random vector and impose a gamma prior on its entries:
\begin{equation}
    \label{eq: gamma prior}
    p(\boldsymbol\lambda^{(k)}) = \prod_{r_k=1}^{R_k} \text{Ga}(\lambda^{(k)}_{r_k}|a_\lambda,b_\lambda).
\end{equation}
The hyperparameters are set to $a_\lambda=1,b_\lambda=5$. 

{\bf Overall Prior for $\boldsymbol{\theta}$.} Combining \eqref{eq:core tensor column prior}, \eqref{eq: first core prior} and \eqref{eq: gamma prior}, we have the overall prior density $p(\boldsymbol{\theta})$:
\begin{equation}
    \label{eq:prior}
    p\left( \boldsymbol{\theta} \right) =p\left(\ten{G}_1\:|\:\boldsymbol{\lambda}^{(1)}\right) p\left(\ten{G}_d\:|\:\boldsymbol{\lambda}^{(d-1)}\right) \prod_{k=2}^{d-1} p\left(\ten{G}_k\:|\:\boldsymbol{\lambda}^{(k-1)},\boldsymbol{\lambda}^{(k)}\right)\prod_{k=1}^{d-1} p(\boldsymbol{\lambda}^{(k)}).
\end{equation}

\subsection{Complete Probabilistic Model}
Now we are ready to obtain the full posterior density by combining \eqref{likelihood},\eqref{eq: posterior} and \eqref{eq:prior}:
\begin{equation}
\label{eq:posterior_full}
\begin{split}
     p(\boldsymbol{\theta} |\ten{D})=& \frac{1}{p(\ten{D})} \prod_{i=1}^N p\left(\mat{y}_i,\mat{g}\left(\mat{x}_i \mid [\![ \ten{G}_1,\dots,\ten{G}_d]\!]  \right) \right) \times \\
     & p\left(\ten{G}_1\:|\:\boldsymbol{\lambda}^{(1)}\right) p\left(\ten{G}_d\:|\:\boldsymbol{\lambda}^{(d-1)}\right) \prod_{k=2}^{d-1} p\left(\ten{G}_k\:|\:\boldsymbol{\lambda}^{(k-1)},\boldsymbol{\lambda}^{(k)}\right)\prod_{k=1}^{d-1} p(\boldsymbol{\lambda}^{(k)}).
\end{split}
\end{equation}
\begin{wrapfigure}{r}{2.4in}
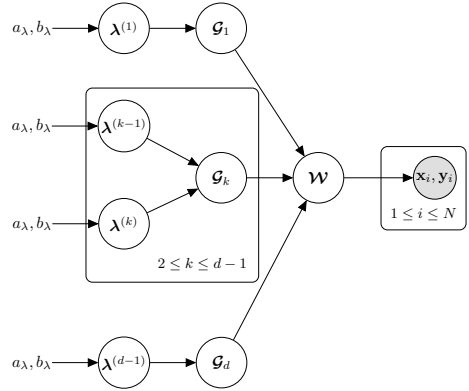

  \centering
\resizebox{\linewidth}{!}{%
  \tikz{
    \node[latent,minimum size = 1.1cm] (lambda1) {$\boldsymbol{\lambda}^{(k)}$} ; %
    \node[latent, above = of lambda1,minimum size = 1.1cm] (lambda2) {$\boldsymbol{\lambda}^{(k-1)}$} ; %
    \node[latent, above = of lambda2,minimum size = 1.1cm] (lambda3) {$\boldsymbol{\lambda}^{(1)}$} ; %
    \node[latent, right=of lambda3, minimum size = 1.1cm] (G1) {$\ten{G}_1$} ; %
    \node[latent, below = of lambda1,yshift = -.9cm,minimum size = 1.1cm] (lambda4) {$\boldsymbol{\lambda}^{(d-1)}$} ; %
    \node[latent, right=of lambda4, minimum size = 1.1cm] (Gd) {$\ten{G}_{d}$} ; %

    \node[const, left = of lambda1] (a) {$a_\lambda,b_\lambda$} ; %
    \node[const, left =of lambda2] (b) {$a_\lambda,b_\lambda$} ; %
    \node[const, left = of lambda4] (c) {$a_\lambda,b_\lambda$} ; %
    \node[const, left = of lambda3] (d) {$a_\lambda,b_\lambda$} ; %

    \node[latent, right=of lambda1, yshift = 1cm, minimum size = 1.1cm] (G) {$\ten{G}_k$} ; %
    \node[latent, right=of G,minimum size = 1.1cm] (w) {$\ten{W}$} ; %
    \plate[inner sep=0.25cm] {plate1} {(lambda1) (lambda2) (G) } {$2\leq k \leq d-1$}; %
    \node[obs, right=of w, xshift = .5cm] (x) {$\mat{x}_i,\mat{y}_i$} ; %
    \plate[inner sep=0.2cm] {plate3} {(x) } {$1\leq i \leq N$}; %
    \edge {a} {lambda1} ; %
    \edge {b} {lambda2} ; %
    \edge {c} {lambda4} ; %
    \edge {d} {lambda3} ; %
    \edge {lambda1} {G} ; %
    \edge {lambda2} {G} ; %
    \edge {lambda3} {G1} ; %
    \edge {lambda4} {Gd} ; %
    \edge {G} {w} ; %
    \edge {G1} {w} ; %
    \edge {Gd} {w} ; %
     \edge {w} {x} ; %
  }
  }
  \caption{Bayesian graphical model for a low-rank Bayesian tensorized neural network parametrized by a single low-rank tensor $\ten{W}$. There are $N$ training samples.}
  \label{fig: graphical model}
\vspace{-30pt}
\end{wrapfigure}


The associated probabilistic graphical model is shown in Fig.~\ref{fig: graphical model}. The user-defined parameters
$a_{\boldsymbol{\lambda}}$ and $b_{\boldsymbol{\lambda}}$ generate a Gamma distribution for $\boldsymbol{\lambda}^{(k)}$ which tunes the actual rank of each TT core $\ten{G}_k$ via a Gaussian distribution. The total number of parameters to be inferred for a single-layer tensorized neural network is $\sum_{k=1}^d M_k J_k R_{k-1}R_{k}+\sum_{k=1}^{d-1} R_k$. The extension to the an $L$-layer neural network is straightforward: one just needs to replicate the whole diagram except the training data by $L$ times.

\paragraph{Automatic Rank Determination.} The actual TT rank is determined by both the prior and training data. As shown in \eqref{eq:core tensor column prior} and \eqref{eq: first core prior}, each entry of $\boldsymbol\lambda^{(k)}$ directly controls one sub-tensor of $\ten{G}_k$ and one sub-tensor of $\ten{G}_{k+1}$. If the entry $\boldsymbol\lambda^{(k)}_{r_k}$ is large then elements of subtensors $\ten{G}_k(:,:,:, r_k)$ and $\ten{G}_{k+1}(r_k,:,:,:)$ can vary freely based on the training data. In contrast, if $\boldsymbol\lambda^{(k)}_{r_k}$ is close to zero, then the elements of $\ten{G}_k(:,:,:, r_k)$ and $\ten{G}_{k+1}(r_k, :, :, :)$ are more likely to vanish. 
Let $\bar{\boldsymbol\lambda}^{(k)}$ be the posterior mean of $\boldsymbol\lambda^{(k)}$ decided by both the prior and training data. Then the inferred TT-rank $\hat{\mat{R}}=[1,\hat{R}_1, \hat{R}_2, \dots, \hat{R}_{d-1},1]$ is estimated as the number of nonzero elements in $\bar{\boldsymbol\lambda}^{(k)}$:
\begin{equation}
\label{eq: new rank}
\hat{R}_k = \text{nnz} \left( \bar{\boldsymbol\lambda}^{(k)}  \right)\; \text{for}\; k=1,2,\dots, d-1.  
\end{equation}
In practice, an element of $\bar{\boldsymbol\lambda}^{(k)}$ is regarded as zero if it is below a threshold. Such an automatic rank tuning reduces the actual number of model parameters in a single layer to $\sum_{k=1}^d M_k J_k \hat{R}_{k-1}\hat{R}_{k}$.

\subsection{Training Algorithms}

    

Now we describe how to train our low-rank Bayesian tensorized neural networks. 

{\bf Full Bayesian Training.} Estimating the posterior density $p(\boldsymbol{\theta} |\ten{D})$ by MCMC~\cite{andrieu2003introduction} is prohibitively expensive for large-scale or deep neural networks. Therefore, we employ the Stein variational gradient descent (SVGD) recently developed by~\citet{liu2016stein} which combines the flexibility of MCMC with the speed of variational Bayesian inference. 

The goal is to find a set of particles $\{\boldsymbol{\theta}^i\}_{i=1}^n$ such that $q(\boldsymbol{\theta})=\frac{1}{n}\sum\limits_{i=1}^n k(\boldsymbol{\theta}, \boldsymbol{\theta}^i)$ approximates the true posterior $p(\boldsymbol{\theta}|\ten{D})$. Here $k(\cdot,\cdot)$ is a positive definite kernel, and we use the radial basis function kernel here. The particles can be found by minimizing the Kullback–Leibler divergence between $q(\boldsymbol{\theta})$ and $p\left(\boldsymbol{\theta}|\ten{D})\right)$. The optimal update $\phi(\cdot)$ is derived in~\cite{liu2016stein} and takes the form 
\begin{equation}
\label{eq: stein update}
\begin{split}
&\boldsymbol{\theta}^k \leftarrow \boldsymbol{\theta}^k+\epsilon\phi(\boldsymbol{\theta}^k), \; \text{with}\; \phi(\boldsymbol{\theta}^k) = \frac{1}{n}\sum_{i=1}^n\left[k(\boldsymbol{\theta}^i,\boldsymbol{\theta}^k)\nabla_{\boldsymbol{\theta}^i}\log p(\boldsymbol{\theta}^i|\ten{D})+\nabla_{\boldsymbol{\theta}^i}k(\boldsymbol{\theta}^i,\boldsymbol{\theta}^k) \right]    
\end{split}
\end{equation}
where $\epsilon$ is the step size. The gradient $\nabla_{\boldsymbol{\theta}}\log p(\boldsymbol{\theta}|\ten{D})$ is expressed as
\begin{equation}
    \label{eq:log posterior gradient}
\nabla_{\boldsymbol{\theta}}\log p(\boldsymbol{\theta} |\ten{D}) = \sum_{i=1}^N \sum_{s=1}^S y_{i,s}\frac{ \nabla_{\boldsymbol{\theta}} \left[g_s\left(\mat{x}_i \mid [\![ \ten{G}_1,\dots,\ten{G}_d]\!]  \right)\right]} {g_s\left(\mat{x}_i \mid [\![ \ten{G}_1,\dots,\ten{G}_d]\!]  \right)}+ \nabla_{\boldsymbol{\theta}} \log p(\boldsymbol{\theta})
\end{equation}
for classification. The first term is exactly the gradient of a maximum-likelihood tensorized training, and it is replaced by a stochastic gradient if $N$ is large. The 2nd term caused by our low-rank prior is our only overhead over standard tensorized training~\cite{novikov2015tensorizing}, and does not require forming the full tensor. Let $\boldsymbol{\Lambda}_k =\text{diag}\left(\boldsymbol{\lambda}^{(k-1)}\otimes\boldsymbol{\lambda}^{(k)}\right)$ be a diagonal matrix, and $\text{vec}\left(\ten{G}_k(:,m_k,j_k,:) \right)$ be the column-major vectorization. We provide the gradients of the log-prior below:
\begin{equation}
    \label{eq: prior grad}
    \small
    \begin{split}
        \frac{\partial \log p(\boldsymbol{\theta})}{\partial \text{vec}\left(\ten{G}_k(:,m_k,j_k,:) \right)} &= -\boldsymbol{\Lambda}_k^{-1} \text{vec}\left(\ten{G}_k(:,m_k,j_k,:) \right)   \\
        \frac{\partial \log p(\boldsymbol{\theta})}{\partial {\lambda}^{(k)}_l}=
        &-\frac{1}{2}\sum_{m_{k+1},j_{k+1},r_{k+1}}\left\{ \frac{1}{{\lambda}^{(k)}_l {\lambda}^{(k+1)}_{r_{k+1}}}-\left(\frac{\ten{G}_{k+1}(l,m_{k+1},j_{k+1},r_{k+1})}{{\lambda}^{(k)}_l {\lambda}^{(k+1)}_{r_{k+1}}}\right)^2\right\}\\
        &-\frac{1}{2}\sum_{r_{k-1},m_k,j_k}\left\{ \frac{1}{{\lambda}^{(k-1)}_{r_{k-1}} {\lambda}^{(k)}_l}-\left(\frac{\ten{G}_k(r_{k-1},m_k,j_k,l)}{{\lambda}^{(k-1)}_{r_{k-1}} {\lambda}^{(k)}_l}\right)^2\right\}+ \frac{(a_{\lambda}-1)}{\lambda^{(k)}_l}-b_\lambda.\\
        \nonumber
    \end{split}
    \normalsize
\end{equation}



{\bf MAP Training.} To obtain the MAP estimation we run stochastic gradient descent to minimize the negative log-posterior:
\begin{equation}
    \label{eq:log posterior}
-\log p(\boldsymbol{\theta} |\ten{D}) = -\sum_{i=1}^N \sum_{s=1}^S y_{i,s}\log g_s\left(\mat{x}_i \mid [\![ \ten{G}_1,\dots,\ten{G}_d]\!]  \right)-
    \log p\left( \boldsymbol{\theta} \right)+\log p(\mathcal{D}) \nonumber
\end{equation}
The local maximum achieved by MAP training provides a single non-Bayesian tensorized neural network that has been compressed by rank determination.

{\bf Initialization.} Training deep neural networks requires careful initializations~\citep{glorot2010understanding}. The same is true for tensorized neural networks~\citep{wang2018wide}. A good empirically determined initialization distribution for the full tensor weights is $\mathcal{N}\left(0,\sqrt{\frac{2}{Q}}\right)$ where $Q$ is the total number of parameters in the full tensor. In order to achieve variance $\sqrt{\frac{2}{Q}}$ for a low-rank TT or TT-matrix with ranks $R_1=\dots = R_{d-1} =R$, we initialize the TT core elements using a $\mathcal{N}\left(0,\sigma^2\right)$ with $\sigma^2 = \left(\frac{2}{Q}\right)^{1/2d}R^{1/d-1}$. This initialization provides a correction to the tensor core initialization described in~\cite{wang2018wide} which contains an error.



\section{Numerical Experiments}

\subsection{Experimental Setup}

We test several network structures on the MNIST and CIFAR-10 datasets to demonstrate our method. We use the Adam optimization algorithm to initialize the first particle~\citep{kingma2014adam} and run 5000 iterations of SVGD. For all trainable weights except the low-rank tensors in our proposed model we apply a $\mathcal{N}(0,100)$ prior which acts as a weak regularizer. We refer to our proposed low-rank Bayesian tensorized model as ``{\bf LR-BTNN}", a Bayesian tensorized neural network with a $\mathcal{N}(0,100)$ prior on all weights and convolution kernels as ``{\bf BTNN}", and a Bayesian non-tensorized neural network with a $\mathcal{N}(0,100)$ prior on all parameters as ``{\bf BNN}". We set up three experiments on the two datasets:

\begin{itemize}[leftmargin=*]

\item {\bf MNIST (2 FC Layers).} First we test on the MNIST dataset~\citep{lecun1998mnist} using a classification neural network with two fully-connected layers. The first layer is size $784\times 625$ with ReLU activation, and the second layer is size $625\times 10$ with a softmax activation. Both layers contain a bias parameter. We convert the first layer into a TT-matrix with $(m_1,m_2,m_3,m_4)=(7,4,7,4), (j_1,j_2,j_3,j_4)=(5,5,5,5)$ and the second layer into a TT-matrix with $(m_1,m_2)=(25,25)$ and $(j_1,j_2)=(5,2)$. Both layers have maximum TT-rank $R_1=\dots=R_{d-1}=20$. We initialize by training for 100 epochs on the log-posterior. We use 50 particles to approximate the posterior density. 

\item {\bf CIFAR-10 CNN (4 Conv layers + 2 FC layers).} We test on the CIFAR-10 dataset~\cite{krizhevsky2014cifar} using a simple convolutional model. This CNN model consists of (Conv-128,BN,ReLU), (Conv-128,BN,ReLU), max-pool $3\times3$, (Conv-256,BN,ReLU), (Conv-256,BN,ReLU), max-pool $5\times 5$, fc-$512$, fc-$10$. All convolutions are $3\times 3$ with stride $1$. Following~\cite{garipov2016ultimate} we do not tensorize the first convolutional layer, which contains less than $1\%$ of the parameters. We set the maximum TT-ranks of all layers to $30$ and initialize all methods by training for 120 epochs on the log-posterior. We use 20 particles for the SVGD fully Bayesian estimation. 

 \item {\bf CIFAR-10 ResNet-110 (109 Conv layers + 1 FC layer).} We further test the baseline Keras ResNetv1 structure~\cite{keras_resnet,keras,he2016deep} on the CIFAR-10 dataset. We follow the learning rate schedule in~\cite{keras_resnet}. We do not tensorize the convolutions in the first Resblock (first 36 layers) or the $1\times 1$ convolutions. For all other convolutions we set the maximum TT-rank to 20. We also use 20 particles for the SVGD fully Bayesian estimation. 

\end{itemize}

\begin{table}[t]
  \caption{Results of a standard Bayesian neural network (BNN), a Bayesian tensorized neural network (BTNN) with Gaussian prior, and our low-rank Bayesian tensorized neural network (LR-BTNN). }
  \label{table: mnist results}
  \centering
  \small
  \begin{tabular}{llllll}
  \hline
 Example & Model & SVGD Param $\#$  & MAP Param $\#$ & Test LL  & MAP Accuracy \\
    \midrule
MNIST    & BNN & 24,844,250  & 496,885 & -0.267 & 92.1\%     \\
(w/ 2 FC  & BTNN & 1,361,750   & 27,235 & -0.188 & 97.7\%    \\
layers)   & {\bf LR-BTNN}  & {\bf 181,250 (137$\times$)}  & {\bf 3,625 (137$\times$)} & {\bf -0.106} &{\bf 97.8\%} \\
    \midrule
    \midrule
CIFAR-10  & BNN & 31,388,360  & 1,569,418 & -0.497 & 89.0\% \\
(CNN, w/& BTNN & 13,737,360  &   686,868 & -0.463  & 88.8\%  \\
4 Conv +2 FC)    & {\bf LR-BTNN} & {\bf 1,987,520 (15.8$\times$)}  & {\bf 99,376 (15.8$\times$)} & {\bf -0.471} &  {\bf 87.4\%} \\
        \midrule
    \midrule
CIFAR-10       & BNN & 34,855,240  & 1,742,762 & -0.506 & 92.6\% \\
(ResNet-110, w/ & BTNN & 12,895,800  & 644,790 & -0.521  & 91.1\%  \\
109 Conv+1 FC)   & {\bf LR-BTNN} & {\bf 4,733,020 (7.4$\times$)}  & {\bf 236,651 (7.4$\times$)} & {\bf -0.515} & {\bf 90.4\%} \\
\bottomrule
  \end{tabular}\normalsize


\end{table}

\subsection{Results}

Table~\ref{table: mnist results} shows the overall performances of our LR-BTNN with BNN and BTNN on the three examples. From the table, we can compare the following performance metrics:

\begin{itemize}[leftmargin=*]

\item {\bf Model Size.} The 3rd and 4th columns of Table~\ref{table: mnist results} show the number of model parameters in the full posterior density estimations and in the MAP estimations, respectively. Because of the automatic tensor rank reduction, our LR-BTNN method generates much more compact neural networks. The compression ratio is very high when the networks have FC layers only ($137\times$ reduction over BNN on the MNIST example). The compression ratio becomes smaller when the network has more convolution layers (e.g., $15.8\times$ on the CNN and $7.4\times$ on ResNet-110). Our method outperforms~\cite{garipov2016ultimate,kim2015compression} which compressed the convolution layers typically by $2\times$ to $4\times$.

\item {\bf Prediction Accuracy.} The 5th and 6th columns of Table~\ref{table: mnist results} show the prediction accuracy of our fully Bayesian and MAP estimations, respectively. For the MAP estimation, the accuracy loss of our LR-BTNN model is very small, and our proposed model even achieves the best performance on the MNIST example. For the fully Bayesian model, we measure the probabilistic model accuracy by computing the predictive log likelihood on held-out test data (denoted as ``Test LL" in the table). A model is more accurate if its ``Test LL" is higher~\cite{gal2016dropout}. Our LR-BTNN performs much better than other two methods on the MNIST example, and all models are comparable on the CIFAR-10 tasks despite the fact that our model is much smaller. 

\end{itemize}



\begin{figure}[t]
\begin{minipage}{.32\linewidth}
    \centering
    \includegraphics[width = \textwidth]{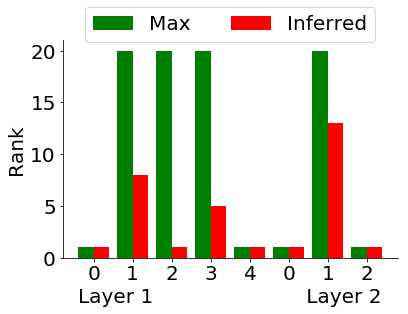}
    (a) {\small Two FC layers in the MNIST example. \normalsize}
\end{minipage}
\begin{minipage}{.32\linewidth}
    \centering
    \includegraphics[width = \textwidth]{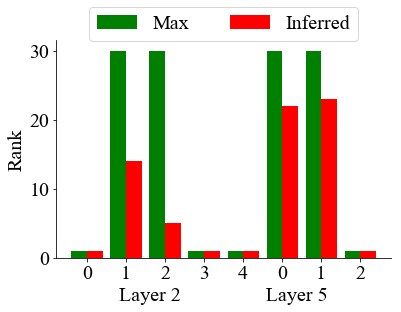}
    (b) {\small Conv (left) and FC (right) layers in CIFAR-10 CNN.\normalsize}
\end{minipage}
\begin{minipage}{.32\linewidth}
    \centering
    \includegraphics[width = \textwidth]{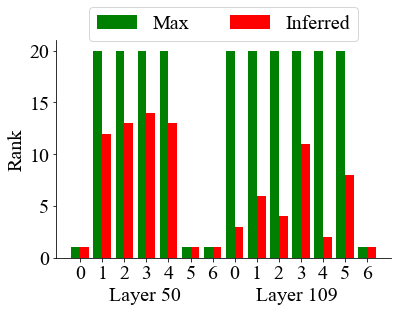}
    (c) {\small Two Conv layers in CIFAR-10 ResNet-110. \normalsize}
\end{minipage}
    \caption{Inferred TT-ranks at some layers in the three examples.}
    \label{fig:rank differences}
\end{figure}

\begin{figure}[t]
	\centering
	\begin{minipage}{.32\linewidth}
	\centering
	    \includegraphics[width = \linewidth]{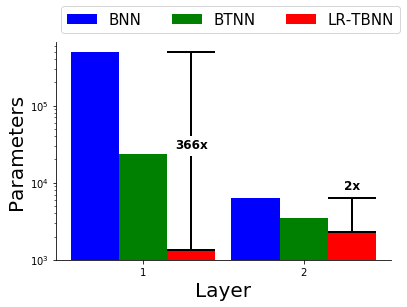}
	    
	    (a) {\small MNIST.\normalsize}
	    
	\end{minipage}
		\begin{minipage}{.32\linewidth}
		\centering
	    \includegraphics[width=\linewidth]{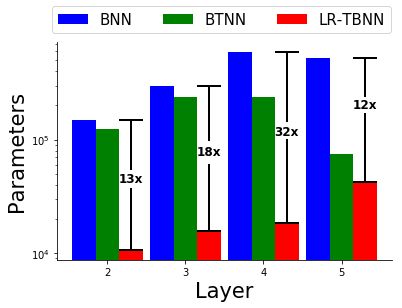}
	    
	    (b) {\small CIFAR-10 CNN.\normalsize}
	\end{minipage}
	\begin{minipage}{.32\linewidth}
	\centering
		\includegraphics[width=\linewidth]{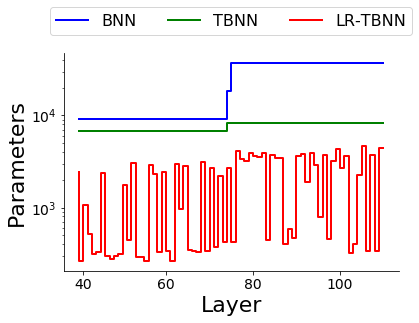}
		
		(c) {\small CIFAR-10 ResNet-110.\normalsize}
	\end{minipage}
\caption{Compression of parameters at different layers due to the automatic rank determination. }
\label{fig: layerwise compression}
\vspace{-10pt}
\end{figure}

{\bf Automatic Rank Determination.} The main advantage of our LR-BTNN method is its automatic rank determination and model compression in the traning process. Fig.~\ref{fig:rank differences} shows the rank determinations of some specific layers in all three examples. Fig.~\ref{fig: layerwise compression} shows the resulting layer-wise parameter reduction due to the automatic rank determination. The MNIST classification model has two overly parameterized FC layers, so the actual TT ranks are much lower than the maximum ranks.
In fact, the TT-rank in Layer 1 reduces from $(1,20,20,20,1)$ to $(1,8,1,5,1)$, and the TT-rank in Layer 2 reduces from $(1,20,1)$ to $(1,13,1)$. This provides $8.5\times$ compression over the naive TNN and overall $138\times$ compression over the non-tensorized neural networks.  A naive tensorization of the CIFAR-10 CNN model gives a compression ratio of $2.3\times$, and the rank determination gives a further parameter reduction of $6.9\times$, leading to an overall $15.8\times$ compression. Most layers in ResNet-110 are convolutions blocks with small numbers of filters, and there is only one small dense matrix. These facts make tensor compression less effective on this ResNet-110 example. 

\begin{wrapfigure}{r}{0.56\textwidth}
    \centering
    \begin{minipage}{0.27\textwidth}
        \centering
           \includegraphics[width = \textwidth, height=0.8\textwidth]{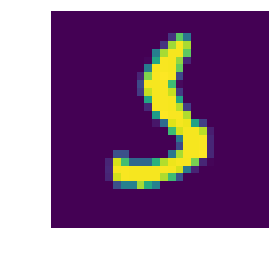} 
    \end{minipage}
    \begin{minipage}{0.27\textwidth}
        \centering
            \includegraphics[width = \textwidth, height=0.8\textwidth]{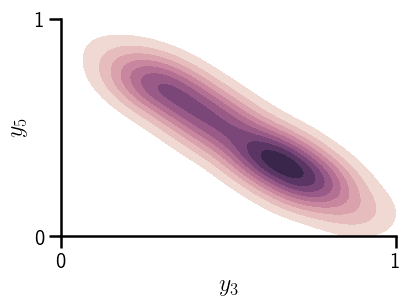}
    \end{minipage}
        \caption{Left: a challenging input image with true label 3. Right:  the joint marginal density of softmax output 3 (x-axis) and softmax output 5 (y-axis).} \label{fig:mnist_uncertainty}
        \vspace{-20pt}
\end{wrapfigure}
\paragraph{Uncertainty Quantification.} Our LR-BTNN method is able to quantify the model uncertainty, which is critical for decision making in uncertain environments. Specifically, based on the posterior density obtained from SVGD, we can easily estimate the probability density of an predicted output. As an example, Fig.~\ref{fig:mnist_uncertainty} shows an image that is hard to classify. The predicted density shows that this image can be recognized as ``3" with the highest probability, but it can also be recognized as ``5" with a high probability.


\section{Conclusion}

We have proposed a low-rank Bayesian tensorized neural networks in the tensor train format. Our formulation provides an automatic rank determination and model compression in the end-to-end training. A Stein variational inference method has been employed to perform full Bayesian estimations, and the resulting model can predict output uncertainties. Our methods have shown excellent accuracy and model compression ratios on various neural network structures with both fully connected and convolution layers for the MNIST and CIFAR-10 data sets. Our method is shown to be more effective on FC layers than on convolution layers.



\newpage
\small
\bibliographystyle{plainnat} 
\bibliography{references.bib}

\begin{thebibliography}{46}
\providecommand{\natexlab}[1]{#1}
\providecommand{\url}[1]{\texttt{#1}}
\expandafter\ifx\csname urlstyle\endcsname\relax
  \providecommand{\doi}[1]{doi: #1}\else
  \providecommand{\doi}{doi: \begingroup \urlstyle{rm}\Url}\fi

\bibitem[ker()]{keras_resnet}
{CIFAR-10 Keras ResNet}.
\newblock \url{https://keras.io/examples/cifar10_resnet/}.

\bibitem[Alvarez and Salzmann(2017)]{alvarez2017compression}
Jose~M Alvarez and Mathieu Salzmann.
\newblock Compression-aware training of deep networks.
\newblock In \emph{Advances in Neural Information Processing Systems}, pages
  856--867, 2017.

\bibitem[Andrieu et~al.(2003)Andrieu, De~Freitas, Doucet, and
  Jordan]{andrieu2003introduction}
Christophe Andrieu, Nando De~Freitas, Arnaud Doucet, and Michael~I Jordan.
\newblock An introduction to {MCMC} for machine learning.
\newblock \emph{Machine learning}, 50\penalty0 (1-2):\penalty0 5--43, 2003.

\bibitem[Bazerque et~al.(2013)Bazerque, Mateos, and
  Giannakis]{bazerque2013rank}
Juan~Andr{\'e}s Bazerque, Gonzalo Mateos, and Georgios~B Giannakis.
\newblock Rank regularization and bayesian inference for tensor completion and
  extrapolation.
\newblock \emph{IEEE transactions on signal processing}, 61\penalty0
  (22):\penalty0 5689--5703, 2013.

\bibitem[Calvi et~al.(2019)Calvi, Moniri, Mahfouz, Yu, Zhao, and
  Mandic]{calvi2019tucker}
Giuseppe~G Calvi, Ahmad Moniri, Mahmoud Mahfouz, Zeyang Yu, Qibin Zhao, and
  Danilo~P Mandic.
\newblock Tucker tensor layer in fully connected neural networks.
\newblock \emph{arXiv preprint arXiv:1903.06133}, 2019.

\bibitem[Carroll and Chang(1970)]{carroll1970analysis}
J~Douglas Carroll and Jih-Jie Chang.
\newblock Analysis of individual differences in multidimensional scaling via an
  {N-way} generalization of “{Eckart-Young}” decomposition.
\newblock \emph{Psychometrika}, 35\penalty0 (3):\penalty0 283--319, 1970.

\bibitem[Chollet et~al.(2015)]{keras}
Fran\c{c}ois Chollet et~al.
\newblock Keras.
\newblock \url{https://github.com/fchollet/keras}, 2015.

\bibitem[Dolgov and Savostyanov(2014)]{dolgov2014alternating}
Sergey~V Dolgov and Dmitry~V Savostyanov.
\newblock Alternating minimal energy methods for linear systems in higher
  dimensions.
\newblock \emph{SIAM Journal on Scientific Computing}, 36\penalty0
  (5):\penalty0 A2248--A2271, 2014.

\bibitem[Gal and Ghahramani(2016)]{gal2016dropout}
Yarin Gal and Zoubin Ghahramani.
\newblock Dropout as a {Bayesian} approximation: Representing model uncertainty
  in deep learning.
\newblock In \emph{International Conference on Machine Learning}, pages
  1050--1059, 2016.

\bibitem[Gandy et~al.(2011)Gandy, Recht, and Yamada]{gandy2011tensor}
Silvia Gandy, Benjamin Recht, and Isao Yamada.
\newblock Tensor completion and low-n-rank tensor recovery via convex
  optimization.
\newblock \emph{Inverse Problems}, 27\penalty0 (2):\penalty0 025010, 2011.

\bibitem[Garipov et~al.(2016)Garipov, Podoprikhin, Novikov, and
  Vetrov]{garipov2016ultimate}
Timur Garipov, Dmitry Podoprikhin, Alexander Novikov, and Dmitry Vetrov.
\newblock Ultimate tensorization: compressing convolutional and {FC} layers
  alike.
\newblock \emph{arXiv preprint arXiv:1611.03214}, 2016.

\bibitem[Glorot and Bengio(2010)]{glorot2010understanding}
Xavier Glorot and Yoshua Bengio.
\newblock Understanding the difficulty of training deep feedforward neural
  networks.
\newblock In \emph{Proc. International Conference on Artificial Intelligence
  and Statistics}, pages 249--256, 2010.

\bibitem[Guhaniyogi et~al.(2017)Guhaniyogi, Qamar, and
  Dunson]{guhaniyogi2017bayesian}
Rajarshi Guhaniyogi, Shaan Qamar, and David~B Dunson.
\newblock Bayesian tensor regression.
\newblock \emph{The Journal of Machine Learning Research}, 18\penalty0
  (1):\penalty0 2733--2763, 2017.

\bibitem[Han et~al.(2015)Han, Mao, and Dally]{han2015deep}
Song Han, Huizi Mao, and William~J Dally.
\newblock Deep compression: Compressing deep neural networks with pruning,
  trained quantization and {Huffman} coding.
\newblock \emph{arXiv preprint arXiv:1510.00149}, 2015.

\bibitem[Hanson and Pratt(1989)]{hanson1989comparing}
Stephen~Jos{\'e} Hanson and Lorien~Y Pratt.
\newblock Comparing biases for minimal network construction with
  back-propagation.
\newblock In \emph{Advances in neural information processing systems}, pages
  177--185, 1989.

\bibitem[Harshman et~al.(1994)Harshman, Lundy, et~al.]{harshman1994parafac}
Richard~A Harshman, Margaret~E Lundy, et~al.
\newblock {PARAFAC}: Parallel factor analysis.
\newblock \emph{Computational Statistics and Data Analysis}, 18\penalty0
  (1):\penalty0 39--72, 1994.

\bibitem[Hawkins and Zhang(2018)]{hawkins2018robust}
Cole Hawkins and Zheng Zhang.
\newblock Robust factorization and completion of streaming tensor data via
  variational bayesian inference.
\newblock \emph{arXiv preprint arXiv:1809.01265}, 2018.

\bibitem[He et~al.(2016)He, Zhang, Ren, and Sun]{he2016deep}
Kaiming He, Xiangyu Zhang, Shaoqing Ren, and Jian Sun.
\newblock Deep residual learning for image recognition.
\newblock In \emph{Proceedings of the IEEE conference on computer vision and
  pattern recognition}, pages 770--778, 2016.

\bibitem[Hillar and Lim(2013)]{hillar2013most}
Christopher~J Hillar and Lek-Heng Lim.
\newblock Most tensor problems are {NP}-hard.
\newblock \emph{Journal of the ACM (JACM)}, 60\penalty0 (6):\penalty0 45, 2013.

\bibitem[Hinton and Van~Camp(1993)]{hinton1993keeping}
Geoffrey Hinton and Drew Van~Camp.
\newblock Keeping neural networks simple by minimizing the description length
  of the weights.
\newblock In \emph{Proc. ACM Conf. on Computational Learning Theory}. Citeseer,
  1993.

\bibitem[Holtz et~al.(2012)Holtz, Rohwedder, and
  Schneider]{holtz2012alternating}
Sebastian Holtz, Thorsten Rohwedder, and Reinhold Schneider.
\newblock The alternating linear scheme for tensor optimization in the tensor
  train format.
\newblock \emph{SIAM Journal on Scientific Computing}, 34\penalty0
  (2):\penalty0 A683--A713, 2012.

\bibitem[Imaizumi et~al.(2017)Imaizumi, Maehara, and
  Hayashi]{imaizumi2017tensor}
Masaaki Imaizumi, Takanori Maehara, and Kohei Hayashi.
\newblock On tensor train rank minimization: Statistical efficiency and
  scalable algorithm.
\newblock In \emph{Advances in Neural Information Processing Systems}, pages
  3930--3939, 2017.

\bibitem[Kendall and Gal(2017)]{kendall2017uncertainties}
Alex Kendall and Yarin Gal.
\newblock What uncertainties do we need in {Bayesian} deep learning for
  computer vision?
\newblock In \emph{Advances in neural information processing systems}, pages
  5574--5584, 2017.

\bibitem[Kim et~al.(2015)Kim, Park, Yoo, Choi, Yang, and
  Shin]{kim2015compression}
Yong-Deok Kim, Eunhyeok Park, Sungjoo Yoo, Taelim Choi, Lu~Yang, and Dongjun
  Shin.
\newblock Compression of deep convolutional neural networks for fast and low
  power mobile applications.
\newblock \emph{arXiv preprint arXiv:1511.06530}, 2015.

\bibitem[Kingma and Ba(2014)]{kingma2014adam}
Diederik~P Kingma and Jimmy Ba.
\newblock Adam: A method for stochastic optimization.
\newblock \emph{arXiv preprint arXiv:1412.6980}, 2014.

\bibitem[Kolda and Bader(2009)]{kolda2009tensor}
Tamara~G Kolda and Brett~W Bader.
\newblock Tensor decompositions and applications.
\newblock \emph{SIAM review}, 51\penalty0 (3):\penalty0 455--500, 2009.

\bibitem[Kossaifi et~al.(2017)Kossaifi, Lipton, Khanna, Furlanello, and
  Anandkumar]{kossaifi2017tensor}
Jean Kossaifi, Zachary Lipton, Aran Khanna, Tommaso Furlanello, and Anima
  Anandkumar.
\newblock Tensor regression networks.
\newblock \emph{arXiv}, 2017.

\bibitem[Krizhevsky et~al.(2014)Krizhevsky, Nair, and
  Hinton]{krizhevsky2014cifar}
Alex Krizhevsky, Vinod Nair, and Geoffrey Hinton.
\newblock The {CIFAR-10} dataset.
\newblock \emph{online: http://www. cs. toronto. edu/kriz/cifar. html}, 2014.

\bibitem[Lebedev et~al.(2015)Lebedev, Ganin, Rakhuba, Oseledets, and
  Lempitsky]{lebedev2014speeding}
Vadim Lebedev, Yaroslav Ganin, Maksim Rakhuba, Ivan Oseledets, and Victor
  Lempitsky.
\newblock Speeding-up convolutional neural networks using fine-tuned
  cp-decomposition.
\newblock In \emph{International Conference on Learning Representations}, 2015.

\bibitem[LeCun(1998)]{lecun1998mnist}
Yann LeCun.
\newblock The {MNIST} database of handwritten digits.
\newblock \emph{http://yann. lecun. com/exdb/mnist/}, 1998.

\bibitem[LeCun et~al.(1990)LeCun, Denker, and Solla]{lecun1990optimal}
Yann LeCun, John~S Denker, and Sara~A Solla.
\newblock Optimal brain damage.
\newblock In \emph{Advances in neural information processing systems}, pages
  598--605, 1990.

\bibitem[Liu et~al.(2013)Liu, Musialski, Wonka, and Ye]{liu2013tensor}
Ji~Liu, Przemyslaw Musialski, Peter Wonka, and Jieping Ye.
\newblock Tensor completion for estimating missing values in visual data.
\newblock \emph{IEEE Trans. Pattern Analysis and Machine Intelligence},
  35\penalty0 (1):\penalty0 208--220, 2013.

\bibitem[Liu and Wang(2016)]{liu2016stein}
Qiang Liu and Dilin Wang.
\newblock Stein variational gradient descent: A general purpose {Bayesian}
  inference algorithm.
\newblock In \emph{Proc. Advances In Neural Information Processing Systems},
  pages 2378--2386, 2016.

\bibitem[Louizos et~al.(2017)Louizos, Ullrich, and
  Welling]{louizos2017bayesian}
Christos Louizos, Karen Ullrich, and Max Welling.
\newblock Bayesian compression for deep learning.
\newblock In \emph{Advances in Neural Information Processing Systems}, pages
  3288--3298, 2017.

\bibitem[Lu et~al.(2019)Lu, Feng, Liu, Lin, Yan, et~al.]{lu2019tensor}
Canyi Lu, Jiashi Feng, Wei Liu, Zhouchen Lin, Shuicheng Yan, et~al.
\newblock Tensor robust principal component analysis with a new tensor nuclear
  norm.
\newblock \emph{IEEE Trans. Pattern Analysis and Machine Intelligence}, 2019.

\bibitem[MacKay(1992)]{mackay1992bayesian}
David~JC MacKay.
\newblock \emph{Bayesian methods for adaptive models}.
\newblock PhD thesis, California Institute of Technology, 1992.

\bibitem[Neal(1995)]{neal1995bayesian}
Radford~M Neal.
\newblock \emph{Bayesian learning for neural networks}.
\newblock PhD thesis, Citeseer, 1995.

\bibitem[Novikov et~al.(2015)Novikov, Podoprikhin, Osokin, and
  Vetrov]{novikov2015tensorizing}
Alexander Novikov, Dmitrii Podoprikhin, Anton Osokin, and Dmitry~P Vetrov.
\newblock Tensorizing neural networks.
\newblock In \emph{Advances in Neural Information Processing Systems}, pages
  442--450, 2015.

\bibitem[Oseledets(2011)]{oseledets2011tensor}
Ivan~V Oseledets.
\newblock Tensor-train decomposition.
\newblock \emph{SIAM J. Scientific Computing}, 33\penalty0 (5):\penalty0
  2295--2317, 2011.

\bibitem[Rai et~al.(2014)Rai, Wang, Guo, Chen, Dunson, and
  Carin]{rai2014scalable}
Piyush Rai, Yingjian Wang, Shengbo Guo, Gary Chen, David Dunson, and Lawrence
  Carin.
\newblock Scalable {Bayesian} low-rank decomposition of incomplete multiway
  tensors.
\newblock In \emph{International Conference on Machine Learning}, pages
  1800--1808, 2014.

\bibitem[Recht et~al.(2010)Recht, Fazel, and Parrilo]{recht2010guaranteed}
Benjamin Recht, Maryam Fazel, and Pablo~A Parrilo.
\newblock Guaranteed minimum-rank solutions of linear matrix equations via
  nuclear norm minimization.
\newblock \emph{SIAM review}, 52\penalty0 (3):\penalty0 471--501, 2010.

\bibitem[Ullrich et~al.(2017)Ullrich, Meeds, and Welling]{ullrich2017soft}
Karen Ullrich, Edward Meeds, and Max Welling.
\newblock Soft weight-sharing for neural network compression.
\newblock In \emph{International Conference on Learning Representations}, 2017.

\bibitem[Wang et~al.(2018)Wang, Sun, Eriksson, Wang, and
  Aggarwal]{wang2018wide}
Wenqi Wang, Yifan Sun, Brian Eriksson, Wenlin Wang, and Vaneet Aggarwal.
\newblock Wide compression: Tensor ring nets.
\newblock In \emph{Proc. IEEE Conference on Computer Vision and Pattern
  Recognition}, pages 9329--9338, 2018.

\bibitem[Yu et~al.(2017)Yu, Zheng, Anandkumar, and Yue]{yu2017long}
Rose Yu, Stephan Zheng, Anima Anandkumar, and Yisong Yue.
\newblock Long-term forecasting using tensor-train {RNN}s.
\newblock \emph{arXiv preprint arXiv:1711.00073}, 2017.

\bibitem[Zhao et~al.(2015)Zhao, Zhang, and Cichocki]{zhao2015bayesian}
Qibin Zhao, Liqing Zhang, and Andrzej Cichocki.
\newblock Bayesian {CP} factorization of incomplete tensors with automatic rank
  determination.
\newblock \emph{IEEE Trans. Pattern Analysis and Machine Intelligence},
  37\penalty0 (9):\penalty0 1751--1763, 2015.

\bibitem[Zhao et~al.(2016)Zhao, Zhou, Zhang, Cichocki, and
  Amari]{zhao2016bayesian}
Qibin Zhao, Guoxu Zhou, Liqing Zhang, Andrzej Cichocki, and Shun-Ichi Amari.
\newblock Bayesian robust tensor factorization for incomplete multiway data.
\newblock \emph{IEEE Trans. Neural Networks and Learning Systems}, 27\penalty0
  (4):\penalty0 736--748, 2016.

\end{thebibliography}
\normalsize

\end{document}